\documentclass[fleqn,10pt]{wlscirep}
\usepackage[utf8]{inputenc}
\usepackage[T1]{fontenc}
\title{Learning to Feel Materials from Multisensory Tactile Data via Interpretable Models}

\author[1]{Li Zou}
\author[1*]{Yasemin Vardar}
\affil[1]{Delft University of Technology (TU Delft), Department of Cognitive Robotics, Delft, 2628 CD, The Netherlands}

\affil[*]{y.vardar@tudelft.nl}

\begin{abstract}
Human tactile perception of materials relies on complex multisensory touch cues, yet the relationship between low-level tactile signals and perceptual representations remains poorly understood. This knowledge gap hinders the integration of touch in digital environments and the development of robots capable of human-like tactile perception. Here we present an interpretable computational framework for modeling human material perception and recognition using multisensory touch data. Our framework comprises three interconnected models: Model 1 maps finger–surface interaction features to psychophysical sensory attributes, Model 2 classifies materials based on these perceptual representations, and Model 3 directly classifies materials from tactile features. The results showed that combining information from pressing, static contact, and sliding interactions improves prediction accuracy, and that thermal cues are particularly informative for both perceptual modeling and material classification. These findings highlight the importance of thermal and compliance cues, which remain underrepresented in current robotic fingers and haptic displays. Incorporating such cues may enhance artificial systems' ability to approximate human material perception and guide the design of more perceptually grounded haptic interfaces.

\end{abstract}
\begin{document}

\flushbottom
\maketitle
\thispagestyle{empty}

\section*{Introduction}

\begin{figure}[t!]
\centering
\includegraphics[width=1\textwidth]{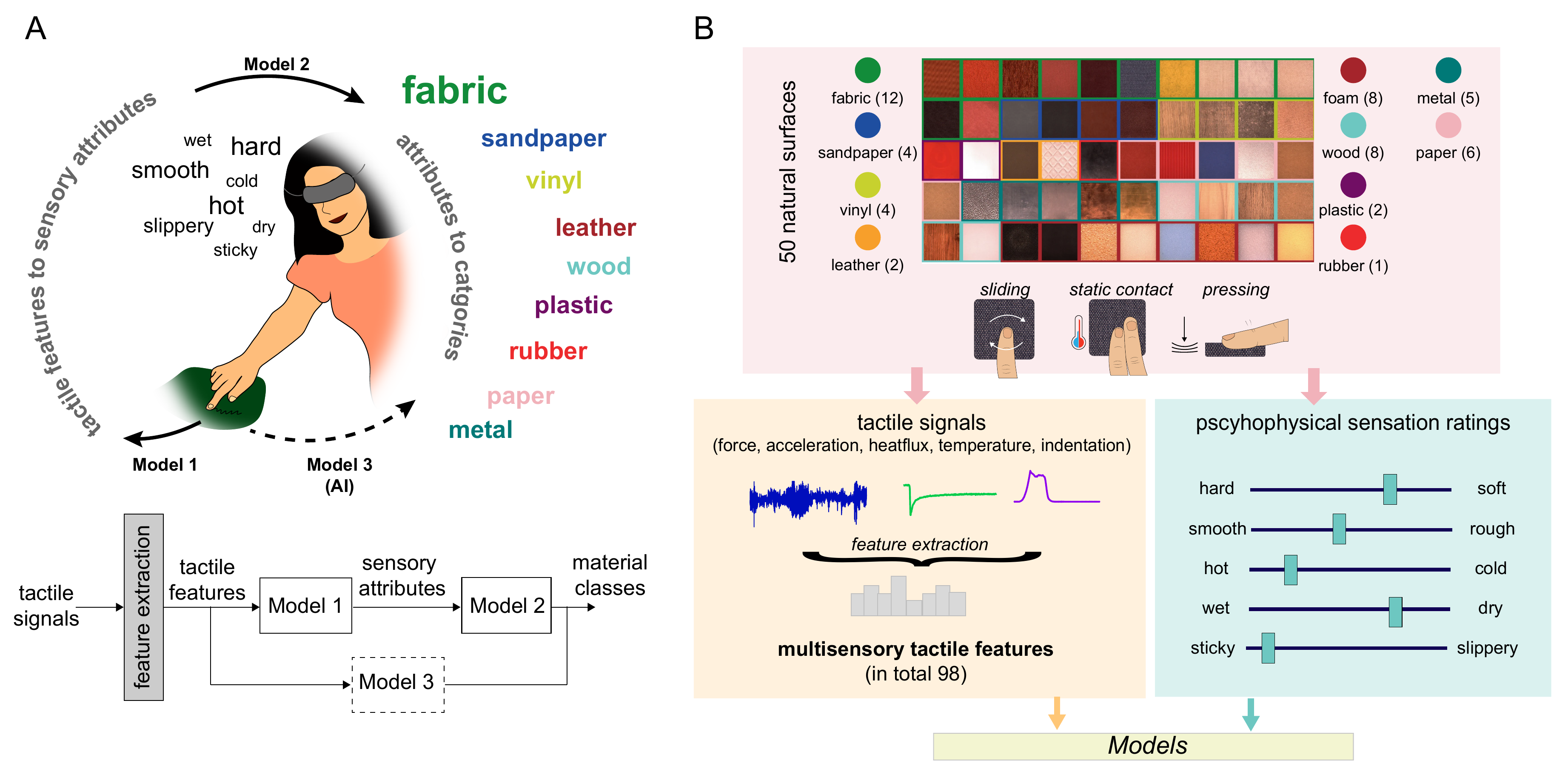}
\caption{Conceptual schematic of the model framework and data preparation. A. Human material recognition and modeling approach. Humans perceive multisensory tactile signals that convey surface attributes, which are integrated to recognize materials. Our framework approximates this process using three models: Model 1 maps extracted features from tactile signals to psychophysical ratings; Model 2 translates these perceptual attributes into material classifications; and Model 3 directly associates tactile features with material categories, serving as a baseline for artificial intelligence classification.
B. Data preparation. Tactile signals and psychophysical sensation ratings were collected from participants interacting with natural surfaces in the SENS3 database~\cite{balasubramanian2025_sens3} using sliding, static contact, and pressing exploratory procedures.}
\label{fig:1}
\end{figure}

Touch plays a fundamental role in how humans perceive and interact with their surroundings, from routine object manipulation—such as holding a glass of water or a toothbrush—to deliberate exploration of surfaces when evaluating materials. Each contact generates a rich array of tactile signals at the skin, which the nervous system interprets to infer sensory attributes, such as roughness, slipperiness, hardness, and warmth, thereby enabling robust material recognition (Figure~\ref{fig:1}). These signals—arising from various exploratory procedures, including pressing, sliding, and static contact, and provide complementary information about surface properties, encompassing both mechanical characteristics (roughness, friction, compliance) and thermal properties~\cite{klatzky2025_action, lederman1987_hand, callier2015_kinematics}. 

Despite humans' remarkable ability to integrate these signals, the processes through which tactile information is transformed into meaningful perceptual attributes and subsequently into material recognition remain incompletely understood. In particular, it remains unclear which sensory cues are most informative, how their combination shapes perception, and how they give rise to distinct sensory attributes. This limited understanding constrains the development of digital interfaces that faithfully replicate touch and hinders the design of robotic systems with human-like tactile perception.

A common approach to investigating these questions relies on psychophysical experiments in which participants rate materials along predefined sensory attributes or categorize them according to material type. The resulting data are typically analyzed using dimensionality reduction techniques to construct a perceptual space. Evidence suggests that the tactile perception of materials can be described by approximately three to five principal perceptual dimensions, commonly associated with rough--smooth, soft--hard, sticky--slippery, and cold--warm attributes~\cite{okamoto2013_dimensions, drewing2018_systematic, hollins1993_multidimensional}. To better understand how these perceptual dimensions arise from physical interactions between the finger and a material, previous studies have primarily examined correlations between the derived perceptual space and measured physical properties of the materials~\cite{BERGMANNTIEST2006_analysis, skedung2020_finishing, vardar219_fingertip, yoshioka2009_direct}. While this approach has provided valuable insights, it often treats individual physical features independently and may overlook complex relationships among multiple sensory cues.

Machine learning has recently emerged as a powerful tool for analyzing complex datasets and has been widely applied to material classification. Early work relied on supervised learning with manually engineered features extracted from tactile or multimodal signals. Although these approaches require domain knowledge to design appropriate feature representations, they remain widely popular due to their relative simplicity and interpretability. Using such methods, classification accuracies ranging from 74\% to 95\% were reported, particularly when multisensory tactile signals (e.g., vibration and deformation) or additional sensory modalities (e.g., visual or auditory data) are incorporated~\cite{Strese2020_exploratory, Devillard2025_database, Zackory2017_semisupervised, Shuvo2025_classification, fishel2012_bayesien}. More recently, deep learning methods~\cite{saga2020_classification, Zhang2024_visuotactile, Zheng2016_deeplearning} have enabled feature learning directly from raw data and achieved high classification accuracies as high as $98.8\%$. 

Beyond material classification, machine learning has been also used to link tactile signals or derived features with perceptual outcomes, such as predicting perceived similarity between surfaces~\cite{richardson2020_learningtopredict, priyadarshini2019_perceptnet} or estimating sensory attributes from tactile signals~\cite{lim2025_machine, richardson2020_learningtopredict}. In addition, perceptual attributes themselves have been used to classify material types, providing insight into how humans categorize materials based on perceptual descriptions~\cite{Baum2013_visual}.

Despite these advances, a comprehensive understanding of how multisensory tactile signals, sensory attributes, and material categories are jointly related within a unified framework remains lacking. Most studies incorporating multisensory tactile signals have focused on improving classification performance~\cite{Strese2020_exploratory, Strese2017_multimodal}, whereas studies addressing human perception have typically examined only a limited subset of signals (e.g., friction or vibration)~\cite{richardson2020_learningtopredict, richardson2022_learn2feel, lim2025_machine} or focused solely on relationships between perceptual attributes and material types~\cite{Baum2013_visual}.

Here, we present an interpretable computational framework to model human material perception and recognition using multisensory touch signals elicited between the finger and surfaces during sliding, pressing, and static contact exploratory conditions. We develop three interconnected models: Model 1 predicts psychophysical sensory attributes from low-level finger–surface interaction signals; Model 2 classifies materials based on these perceptual attributes, approximating human recognition; and Model 3 directly maps tactile features to material categories, serving as a baseline for artificial intelligence–based classification. This hierarchical framework enables systematic analysis of how multisensory tactile signals are transformed into perceptual representations and material recognition. Our results highlight the relative contributions of different sensory modalities and exploratory procedures, emphasizing the critical role of thermal and compliance cues. These findings advance understanding of human tactile perception and inform the design of robots capable of more human-like material recognition, as well as haptic interfaces that can generate realistic virtual sensations.

\section*{Results}

\subsection*{Modeling framework and data preparation}
To evaluate material perception and identification through a human-mimetic approach, we developed three distinct computational frameworks that simulate different stages of the tactile recognition pipeline (Fig.~\ref{fig:1}A). Model 1 maps tactile signal features to psychophysical sensations, while Model 2 utilizes these sensations to predict material categories. Together, these models represent a mediated recognition path. In contrast, Model 3 serves as a direct classification benchmark, mapping tactile features to materials without intermediate perceptual anchors. This comparative architecture allows us to contrast an "interpretable" decision-making process—constrained by human sensory dimensions—against the unguided, high-dimensional performance of standard AI.

Tactile data were sourced from SENS3, an open-access multisensory dataset containing recordings of fingertip interaction signals with 50 distinct surfaces, spanned across 10 different materials---metal, wood, fabric, paper, rubber, plastic, sandpaper, leather, foam, vinyl ~\cite{balasubramanian2025_sens3} (Fig.~\ref{fig:1}B). These signals were captured as participants performed various exploratory procedures—specifically static contact, pressing, and sliding. Signal features (a total of 98) were extracted for each procedure and selectively introduced into the models. Additionally, new psychophysical experiments were conducted to obtain sensation ratings of the materials using five pairs of adjectives: rough-smooth, sticky-slippery, hot-cold, hard-soft, and wet-dry. Comprehensive information regarding the acquisition protocol and feature extraction are provided in the Methods. 

To achieve the best predictive performance, we evaluated a comprehensive set of supervised learning algorithms, ranging from traditional estimators such as K-Nearest Neighbors~\cite{Fix+Hodges:1951, Cover1967} and Support Vector Machines (SVMs)~\cite{SVM1, PhysRevE.111.045306} to ensemble methods, such as Random Forests~\cite{RF, Ho1998} and Gradient Boosting~\cite{Hastie2009,  GB}, as well as Multi-layer Perceptrons (MLPs)~\cite{Cybenko1989, NN}. Each final model was selected based on its peak regression or classification accuracy for its specific role in the hierarchy. 

\subsection*{Mapping tactile features to perceptual attributes (Model 1)}

This model establishes a mapping from low-level tactile features to psychophysical sensation ratings, which represent as higher-level sensory attributes. Using a Random Forest regressor, we assessed how effectively tactile signal–derived features predict human sensation ratings and identified the features that contribute most strongly to each adjective pairs.

The modeling approach was two-fold: first, a comprehensive model utilized the complete feature set---encompassing pressing, static contact (thermal), and sliding data---to establish an upper bound for predictive performance. Second, we conducted a feature-group ablation study by training independent models on each feature set. This procedure allowed us to quantify the specific contribution of each individual exploratory procedures to the prediction of discrete perceptual dimensions. Model performance was quantified using Mean Squared Error (MSE) and the coefficient of determination ($R^2$), where a minimized MSE and an $R^2$ approaching 1 denote superior predictive accuracy. The results were bench marked against a null model that generated predictions by sampling from a uniform distribution $U(1, 0)$.

\begin{figure}[ht!]
\centering
\includegraphics[width=1\textwidth]{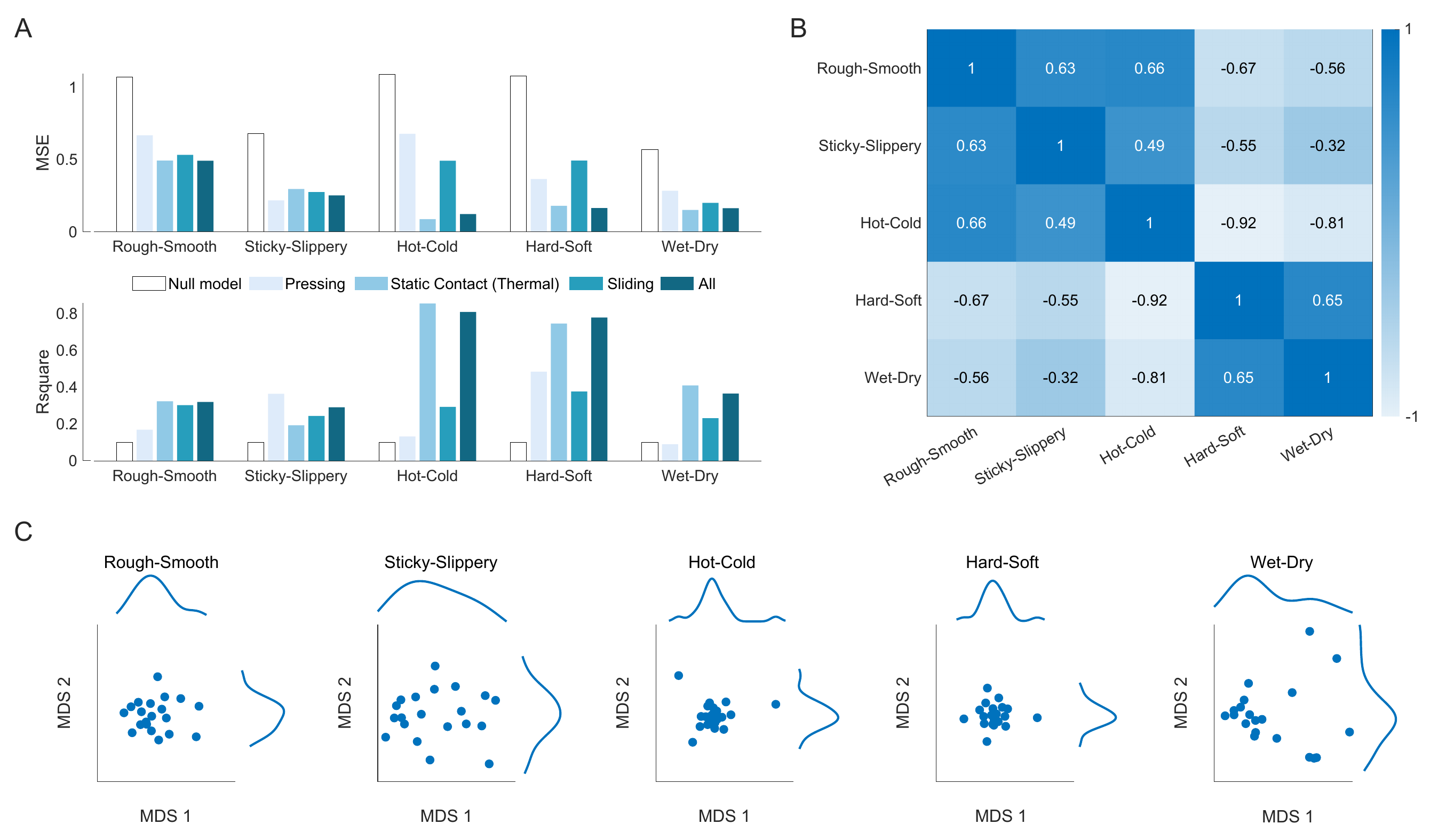}
\caption{Results for interpreting the initial stage of human tactile processing, in which features are mapped to perceptual attributes (Model 1). (A) The performance of the regression models, evaluated using Mean Squared Error (MSE) and R-squared ($R^2$) scores. A lower MSE and a higher $R^2$ indicate better regression performance. All results were obtained using Random Forest regression, except for the null model, which serves as a baseline. (B) Spearman correlation between all pairs of sensation ratings. (C) Illustration of multidimensional scaling (MDS) for adjective pairs. In each subplot, each dot represents a participant, and the distance between any two dots reflects the difference in their ratings across the 50 textures. }
\label{fig:2}
\end{figure}

Across all sensation ratings, the regressor outperformed the null model, even in cases where predictive accuracy was modest (Fig.~\ref{fig:2}A). Feature-specific analysis revealed that static contact data, which captures thermal interaction signals, yielded the best performance; it successfully predicted sensation pairs, such as hot-cold, hard-soft, and wet-dry. While pressing and sliding data also contributed---particularly in predicting sticky-slippery and rough-smooth, respectively---the integrated model (combining all modalities) effectively synthesized these individual strengths to achieve the most robust predictions. Detailed analyses of the contributions of each feature to predicting each sensory attribute are shown in Supplementary Fig.~S1 and Supplementary Table~S1. 

It is intuitive that thermal data can accurately predict the hot-cold~\cite{ho2017_thermalrecognition} and wet-dry sensation\cite{filingeri2014_wetness}. However, it is less obvious why thermal data can also predict the hard-soft sensation. This finding can be explained by the strong correlation between the ratings for the hot–cold and hard-soft adjective pairs (Fig.~\ref{fig:2}B). These results suggest that, in our dataset, materials perceived as warmer (or colder) also tend to be perceived as softer (or harder), respectively. This coupling between sensory attributes accounts for the ability of thermal signals to predict hard–soft judgments in this case. An additional contributing factor may be the physical interaction between the finger and the material surface: softer materials increase the contact area and reduce the air gap at the interface, thereby lowering thermal resistance and altering heat transfer~\cite{bermanntiest2010_tactual}.

Moreover, pressing data appear to predict sticky–slippery sensations more effectively than sliding data. While slipperiness can be readily inferred from friction-related features during sliding, previous study~\cite{nam2020_stickiness} have shown that fingerpad detachment during pressing provides informative cues about stickiness by breaking adhesive bonds at the contact. 

Furthermore, the results revealed that hot-cold and hard-soft sensations were easier to predict compared to rough-smooth, sticky-slippery, and wet-dry sensations. This discrepancy aligns with the inter-participant agreement across adjective pair ratings; participants tended to provide more consistent and similar ratings for hot-cold and hard-soft, whereas ratings for the other sensation pairs showed greater variability (Fig.\ref{fig:2}C). As the predicted sensation ratings in this study are based on average scores, greater individual differences make the averaged ratings less representative, thereby making these sensations more challenging to predict. Additionally, capturing these sensory attributes may require an expanded or more expressive feature space.

These results are consistent with the findings of Richardson and Kuchenbecker~\cite{richardson2020_learningtopredict}, who predicted perceptual distributions of adjective ratings assigned by human participants using sensor data collected with a robotic finger (BioTac) capable of acquiring multimodal tactile signals. Using an unsupervised dictionary-learning approach, they found that prediction errors were lowest for adjectives related to thermal properties (e.g., cold, warm) and compliance (e.g., hard, springy). Interestingly, their algorithm was also able to predict stickiness accurately, but not slipperiness. This discrepancy may be attributable to the elastomeric skin of the robotic finger, which could amplify stickiness-related signal features relative to those produced by the human finger. Alternatively, it may reflect properties of the selected materials, for which human participants provided more consistent ratings, or indicate that additional features are required to reliably capture this attribute.

\subsection*{Mapping sensory attributes to material types (Model 2)}
This model examines material classification based on psychophysical sensation ratings, which reflect higher-level human perceptual attributes. We assessed how effectively these sensory descriptors capture differences among materials and support discrimination across material classes.

\begin{figure}[t!]
\centering
\includegraphics[width=0.99\textwidth]{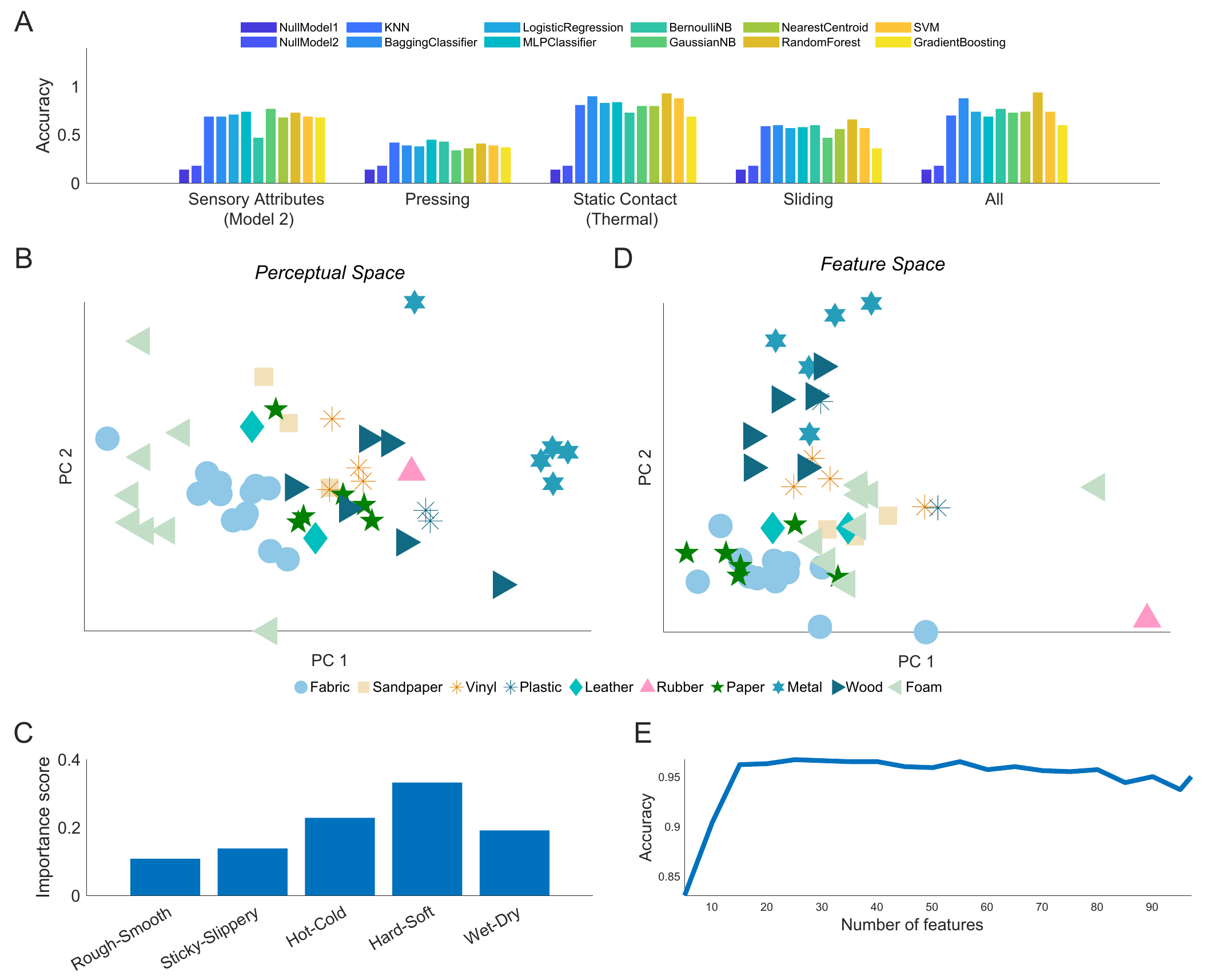}
\caption{Results comparing models of higher-level human cognitive processing (Model 2) and direct AI-based classification (Model 3). 
(A) Classification accuracy for material identification using sensory attributes data, pressing data, static contact, sliding, and a combined dataset (pressing, static contact, and sliding). 
(B) Scatter plot of all materials in two-dimensional perceptual space derived from Principle Component Analysis (PCA), which each point representing a material positioned according to its first and second principle component scores. 
(C) Feature importance scores for all sensation ratings computed using a Random Forest model trained to classify material classes from sensory attributes. 
(D) Scatter plot of all materials in two-dimensional feature space obtained from PCA. 
(E) Classification accuracy as a function of the number of top-ranked features selected by Random Forest, using the combined dataset (pressing, static contact, and sliding) for material classification.}
\label{fig:3}
\end{figure}

The material classification accuracy was quantified using a range of machine learning algorithms. The collected sensation ratings were effective in predicting material types (Fig.~\ref{fig:3}, first column). Among the models tested, Logistic Regression, MLP Classifier, GaussianNB, and Random Forest demonstrated the highest prediction accuracy, all exceeding $70\%$.

The effectiveness of psychophysical sensation data for material classification was further examined using principle component analysis. The results showed that the first two principle components (PC1 and PC2) accounted for over $80\%$ of the total variance, indicating that these components provide a reliable low-dimensional representation for visualizing the materials (Fig.~\ref{fig:3}B). Materials from the same category tended to cluster together in this space, suggesting that the psychophysical sensation data effectively distinguished between different material types.

We next sought to understand the relative importance of each sensory attribute for classification accuracy and to determine whether these attributes contribute independently or interact synergistically. Addressing these questions is particularly important, as the relative weighting of sensory cues may reveal perceptual hierarchies in human haptic processing. For this analysis, we used Random Forest algorithm. The attributes hard-soft and hot-cold showed higher importance scores than wet-dry, sticky-slippery, and rough-smooth (Figure~\ref{fig:3}C). Moreover, the hard–soft dimension emerged as the most influential factor, suggesting its perceptual primacy in material discrimination.

The material classification accuracy obtained from sensation ratings, together with the relative importance of these ratings, shows strong agreement with the findings of Baumgartner et al.~\cite{Baum2013_visual} In their study, human participants were asked to categorize 84 surfaces spanning seven material categories using haptic cues alone and achieved a classification accuracy of 66\%. Their results indicated that successful classification was primarily driven by adjective ratings related to compliance and thermal properties of the surfaces, closely mirroring the patterns observed in our study. Taken together, these findings suggest that sensation ratings can serve as an effective input for achieving human-level material categorization, supporting the role of higher-level human perceptual representations in material identification.

\subsection*{Mapping tactile features to material types (Model 3)}
This model focuses on material classification using features extracted directly from tactile signals. We assessed whether these features provide an efficient and representative encoding of material properties and identified which data types and individual features contribute most strongly to classification accuracy. 

We evaluated classification accuracy using a range of machine learning algorithms. The results showed that the machine learning models performed remarkably well in classifying materials using tactile features (Fig.~\ref{fig:3}A, second to last column). When the tactile features extracted from pressing, static contact (thermal), and sliding data are used independently, thermal data consistently yielded the highest accuracy across all machine learning models (Fig.~\ref{fig:3}A, third column). At the same time, pressing data achieved lowest accuracy (Fig.~\ref{fig:3}A, second column). When using the combined dataset (pressing, static contact, and sliding) for material classification, an accuracy of $94\%$  was achieved using the Random Forest model (Fig.~\ref{fig:3}A, last column). Among the models evaluated, Random Forest algorithm achieved the best overall classification accuracy. 

To better understand the underlying mechanisms of our results, we conducted a principle component analysis on the feature space. The first and second principal components together explained around $40\%$ of the total variance, indicating that additional dimensions are likely needed for more accurate material classification based on tactile data. However, even for two-dimensional space (Fig.~\ref{fig:3}D), certain materials-such as wood, metal, and sandpaper-already tend to form their own clusters. 

Since the Random Forest algorithm achieved the best overall classification accuracy, we used it to analyze feature importance. Each of the 98 features was assigned an importance score based on its contribution to material classification, as estimated by the Random Forest algorithm. Classification accuracy was then evaluated as a function of the number of top-ranked features using the combined dataset (pressing, static contact, and sliding). Using only the top five features, all derived from thermal data (Supplementary Fig.~S1), resulted in over $80\%$ accuracy (Fig.~\ref{fig:3}E). The maximum accuracy of $96.7\%$ was achieved using the top 25 features. Beyond this point, accuracy slightly decreased, likely due to the inclusion of less informative or redundant features. Detailed analyses of the contributions of each feature to material classification are shown in Supplementary Fig.~S1 and Supplementary Table~S1. 

Our results demonstrate higher classification accuracy than prior work that relied solely on friction-induced vibration data to classify ten different surfaces~\cite{Devillard2025_database}, highlighting the importance of incorporating multisensory tactile information for material classification. Moreover, our achieved accuracy is comparable to the results reported by Strese et al.~\cite{Strese2020_exploratory}, who used multisensory tactile data collected from surfaces to classify 184 materials.

\section*{Discussion}

In this work, we present an interpretable computational framework for modeling human material perception and recognition using multisensory touch data. We developed three interconnected models that progressively map tactile information across representational levels, from low-level interaction signals to perceptual attributes and material categories. Model 1 captures the relationship between tactile features derived from finger–surface interactions and psychophysical sensory attributes reported by human participants. Model 2 uses these perceptual representations to classify materials in a manner that approximates human recognition, while Model 3 directly classifies material types from tactile features using machine learning. Together, these models enable a systematic examination of how different representations contribute to material recognition without making claims about the underlying biological or neural mechanisms of human perception.

The interpretable modeling approach adopted here provides a flexible framework for analyzing and comparing representations of tactile information across human and artificial systems. Rather than aiming to explain the underlying neural mechanisms of perception, this framework provides a practical tool for evaluating how sensory cues and exploratory strategies contribute to perceptual representations and recognition performance. As such, it may inform the design of future tactile sensing systems and haptic interfaces that are more closely aligned with human perceptual behavior.

Our results demonstrate that multisensory tactile data obtained through diverse exploratory procedures play a critical role in modeling material perception. Integrating information from pressing, static contact, and sliding interactions consistently improved predictive performance, highlighting the complementary contributions of different exploratory behaviors. Among the sensory modalities examined, thermal cues emerged as particularly informative for both perceptual modeling and material classification. This finding aligns with prior works on human haptic perception~\cite{Baum2013_visual, okamoto2013_dimensions} and suggests that thermal information provides a salient and robust signal for differentiating materials within computational models.

Despite their importance in perceptual modeling, thermal cues remain underrepresented in current robotic fingers and haptic display technologies, which predominantly focus on single modality mechanical sensing~\cite{lepora} and actuation~\cite{chen2026_review, biswas2019_emerging}. Our results suggest that incorporating sensors capable of encoding thermal information may improve robotic systems' ability to approximate human-like material perception. Similarly, haptic interfaces could benefit from richer multisensory display capabilities, including thermal feedback, to better approximate real-world material interactions in virtual and digital environments~\cite{vardar2025_multimodal, Kodak2023}.

\section*{Methods}

\subsection*{Dataset}

For tactile signals, we utilized SENS3, an open-access multisensory dataset collected from fifty different surfaces~\cite{balasubramanian2025_sens3}. Tactile data were recorded using an apparatus equipped with a 3D accelerometer, a 6D force-torque sensor, a thermistor, a heat-flux sensor, and an infrared position sensor. Signals were collected while two participants explored these surfaces with their fingertips using various exploratory actions, including static contact, pressing, and sliding. Further details regarding the data recording procedure can be found in \cite{balasubramanian2025_sens3}.

For perceptual data, psychophysical experiments were conducted with 20 participants, following a procedure similar to Kodak and Vardar\cite{Kodak2023}. Prior to the experiment, participants were informed about the study’s objectives and procedures and provided written informed consent. The protocol was approved by the Human Research Ethics Committee at TU Delft (number 3469) and conducted in accordance with the Declaration of Helsinki.

During the perceptual experiment, participants' task was to explore the surfaces using their dominant index finger with three exploratory motions: static contact, pressing, and sliding. After each interaction, they were asked to rate their perceived sensations using five pairs of adjectives: rough-smooth, sticky-slippery, hot-cold, hard-soft, and wet-dry---which represent the primary psychophysical dimensions of tactile surface perception found in the literature~\cite{okamoto2013_dimensions, Baum2013_visual, balasubramanian2025_sens3}. Ratings were collected using the semantic differential method~\cite{okamoto2013_dimensions} using a 15-point scale (Supplementary Fig.~S2). Participants provided their responses by adjusting sliders within a graphical user interface (GUI) designed in MATLAB, evaluating each surface across five pairs of opposing descriptors. The surfaces were presented in a randomized order. 

The surfaces were placed inside an opaque box to ensure that the perceptual data were based solely on touch. One side of the box was covered with a curtain, allowing participants to put one of their hands inside and explore the surfaces without visual cues. Participants also wore headphones playing white noise to eliminate auditory cues (Supplementary Fig.~S2).  

Before the main experiment, participants were asked to wash and dry their hands. They then completed a familiarization phase, exploring all 50 surfaces while wearing headphones and an eye mask. This phase was followed by mock trials with five randomly selected surfaces to help participants become accustomed to the experimental procedure. Each experimental session lasted approximately one hour.

For a given sensation pair, each participant's rating form a vector of 50 elements, where the $ith$ element corresponds to the rating for the $ith$ surface. To account for the inter-participant variability in scoring tendencies, ratings were normalized to the [0, 1] range for each participant. Then, the resulting normalized ratings were averaged across participants, yielding a 50$\times$8 matrix (surfaces $\times$ adjective pairs). 

\subsection*{Feature extraction}
\label{feature_engineering}
We extracted features from the finger-surface interaction data recorded during three exploratory actions (pressing, static contact, and sliding) to capture the intrinsic properties of each surface. These features served as inputs to machine learning models for both material classification and sensation regression. 
The core approach of our feature extraction process is as follows: for each type of data, we first segmented the original signal into smaller and representative parts  (Supplementary Fig.~S3). We then performed curve fitting on these segments and extracted the parameters of the fitted curves. These parameters capture key characteristics of the tactile signals and were used as features for classification. Table S1 summarizes all the features for each data type, along with the feature importance scores for each model. 

\subsubsection*{\textbf{Pressing}}
The pressing data captures how different surfaces respond to normal pressure applied by the index finger. During data collection, the participant rested their right hand on a support while a linear stage moved vertically, pressing the fingertip against the surface until a normal force of 3~N was reached. This force was maintained for two seconds before the finger was returned to its initial position (Supplementary Fig.~S3).

To characterize the relationship between applied force and indentation depth, we extracted two segments from the raw pressing data, each corresponding to a distinct interaction phase. The pressing phase beings at first contact and continues as the applied force increases until maximum indentation depth is reached. The lifting phase corresponds to the release of force, during which the indentation depth gradually returns toward its baseline.

For both phases, indentation depth was analyzed as a function of applied force. To ensure comparability across surfaces, the indentation depth at initial contact was subtracted from all measurements, normalizing the starting depth to zero. The resulting force-depth relationships capture the compliance of surfaces during pressing and lifting. These curves were subsequently parametrized and used as features for the machine learning models. 
 
As expected, indentation depth increased with applied force, following an exponential trend. To model this relationship, we fit the force-indentation data with the function: $y=a(1-e^{-bx})+c$, where $x$ is the applied normal force and $y$ is the indentation depth.  In this formulation, $a$ represents the maximum indentation depth, $b$ indicates the rate of change, and $c$ corresponds to the initial indentation offset at zero applied force. 

For the pressing stage, the fitted parameters $\{a,b,c\}$ are denoted as $\{a^{P1},b^{P1},c^{P1}\}$. Since indentation depth is normalized to zero at the moment of first contact, $c^p = 0$, allowing us to represent the pressing stage solely by $\{a^{P1},b^{P1}\}$. For the lifting stage, the fitted parameters $\{a,b,c\}$ are denoted as $\{a^{P2},b^{P2},c^{P2}\}$. These parameters collectively capture each surface's compliance and relaxation behavior under loading and unloading. 

To evaluate model accuracy, we computed the coefficient of determination $R^2$, which quantifies the proportion of variance explained by the fit. We also examined the probability density functions of $R^2$ across all surfaces and participants. Both the pressing and lifting stages yielded consistently high $R^2$ values, with distributions concentrated near 1, confirming an excellent goodness of fit.

From this modeling step, the fitted parameters $\{a^{P1},b^{P1},a^{P2},b^{P2},c^{P2}\}$ were extracted as candidate features for subsequent machine learning analyses. In addition, we computed the average difference in indentation depth at the same applied force between the pressing and lifting phases--denoted as $\Delta^{P}$. This hysteresis-based feature reflects the surface's differential response to loading and unloading and serves as an additional discriminative curve. 

\subsubsection*{\textbf{Thermal}}
Thermal data were collected to quantify static contact interactions during finger-surface contact, including heat flux from the skin and corresponding skin temperature.

The heat flux signals were first smoothed using a moving average filter. Then, the features were extracted from two representative segments of this signal. The first segment spans from the onset of heat flux decrease to the minimum, capturing the primary phase of heat transfer. The second segment extends from the minimum to 30 seconds, representing stabilized thermal behavior.

To model the thermal response dynamics, we fitted two functions to consecutive segments. For the first segment, a logistic function $y=\frac{a}{1+e^{-b(x-c)}}$ was used, where $x$ is time, $y$ is heat flux, $a$ represents the asymptotic value as $x\rightarrow\infty$, $b$ is the steepness of the curve, and $c$ denotes the sigmoid midpoint. For the second segment, a power-law function $y=ax^b+c$ was fitted. 

The logistic fit yields parameters $\{a,b,c\}$, denoted as $\{a^{H_1},b^{H_1},c^{H_1}\}$, while the  power-law fit yields parameters $\{a,b,c\}$, denoted as $\{a^{H_2},b^{H_2},c^{H_2}\}$. Collectively, these six parameters, $\{a^{H_1},b^{H_1},c^{H_1},a^{H_2},b^{H_2},c^{H_2}\}$, serve as features that capture surface properties under static contact action.

A similar procedure was applied to the skin temperature data. First, the raw data was smoothed to reduce noise; second, representative segments were identified for curve fitting; and third, the fitted parameters were extracted as features. 
Due to the inherent noise in the signal, we selected only the interval where the temperature begins to drop sharply until it reaches its minimum point for curve fitting. This phase captures the initial change in finger temperature upon contact with the material.

Based on the shape of the temperature data over time during this stage, we propose using a four-parameter logistic (4PL) function to fit the data. The function is defined as: $y=d+\frac{a-d}{1+(\frac{x}{c})^b}$, where we can set $x$ represents time and $y$ represents temperature. The parameters in this function can be interpreted as follows: $a$ is the initial temperature value when $x$ is very small; $b$ is the slope factor, indicating the steepness of the curve; $c$ is the inflection point--the $x$-value where the curve changes direction and the slope is maximal; and $d$ the asymptotic temperature value as $x$ becomes very large. The optimal fitted parameters are denoted as ${a^{T},b^{T},c^{T},d^{T}}$. 
Besides, we also studied the goodness of fit across all materials and participants. We found the $R^2$ values are consistently close to 1, indicating that the proposed function provides an excellent fit for the temperature data within the selected region.

Since the previously selected region represents only a small portion of the entire temperature data, we also extract five representative temperature values from the full time series: (1) The highest temperature before the temperature begins to drop dramatically, denoted as $a^{T0}$. (2) The first minimum temperature recorded after the finger makes contact with the surface, denoted as $b^{T0}$. (3) The highest temperature following the initial minimum, denoted as $c^{T0}$. (4) The temperature at the end of the touch duration, denoted as $d^{T0}$. From the raw temperature data, a local $k$-point median absolute deviation (MAD) was computed, where each MAD is computed over a sliding window of length $k$ across neighboring elements of temperature data. For each sliding window $A$ with $k$ scalar observations, the median absolute deviation (MAD) is defined as $MAD=median (|A_i-median(A)|)$, where $A_i$ is the $ith$ element of $A$ and $i=1,2,...,k$. MAD indicates how spread out a dataset is around its median. 
We observed that the moving median absolute deviation reaches a peak at the point where the fingers touch the material. This peak may serve as a useful feature for capturing temperature changes. For this reason, we included it as a temperature-related feature. These values provide a broader view of the temperature dynamics throughout the entire touch interaction.

\subsubsection*{\textbf{Sliding}}
Sliding data aims to capture the friction and roughness properties of the surfaces. The features from the sliding data were selected based on previous literature~\cite{richardson2022_learn2feel, Strese2017_multimodal, Strese2020_exploratory}  
We first calculated the friction coefficient ($s_{13}$) data by dividing lateral force to normal force calculating the mean of the obtained signal. Afterwards, we applied a bandpass filter (20~Hz--1~kHz) to both lateral force and acceleration data. Then, we extracted features in both the time and frequency domains for the lateral force and acceleration signals. In the time domain, the extracted features include the mean ($s_1$), standard deviation ($s_2$), root mean square ($s_3$), maximum value ($s_4$), skewness ($s_5$), kurtosis ($s_6$), shape factor ($s_7$), impulse factor ($s_8$), crest factor ($s_9$), clearance factor ($s_{10}$), total harmonic distortion ($s_{11}$), signal to noise and distortion ratio (sinad $s_{12}$), and average power ($s_{14}$). These features describe the statistical and physical properties of the force signal during sliding. In the frequency domain, we computed features such as spectral centroid ($s_{15}$), spectral spread ($s_{16}$), spectral roll-off ($s_{17}$), spectral flatness ($s_{18}$), bandwidth ($s_{19}$), peak frequency ($s_{20}$) and total energy across five frequency bands: 0-100 ($s_{21}$), 100-500 ($s_{22}$), 500-1000 ($s_{23}$), 1000-2000 ($s_{24}$), and 2000-5000~Hz ($s_{25}$), respectively. The features corresponding to acceleration signals are denoted as $s_{27}$, $s_{28}$, ..., $s_{49}$.
These features capture the frequency distribution and texture of the signal, offering insight into the dynamic behavior of the material during sliding interactions. 

Moreover, we also used Mel-frequency cepstral coefficients (MFCCs), as they are widely used features in speech signal processing for recognition tasks. Specifically, we extracted the first 14 MFCCs as features for both lateral force an acceleration signals. And they are denoted as $MFCC_{1}$, $MFCC_{2}$, ..., $MFCC_{28}$, respectively.

\subsection*{Model training}
During the machine learning training process, $80\%$ of the data points are used for training the model, while the remaining $20\%$ are reserved for testing. All reported results are obtained from the test data. To minimize the effects of randomness in the data split, this process is repeated 100 times, and the final results are calculated as the average over these iterations.

All machine learning algorithms were implemented using the Python library scikit-learn (sklearn). Supplementary Table~S2 lists the tuned hyperparameters for each algorithm; any parameters not listed were left at their default values as specified by the package.

\bibliography{bibliography}

@article{yoshioka2009_direct,
author = {Yoshioka, T and  Bensmaia, S. J. and Craig, J. C. and S. S. Hsiao},
title = {Texture perception through direct and indirect touch: An analysis of perceptual space for tactile textures in two modes of exploration},
journal = {Somatosensory Motor Research},
volume = {24},
number = {1-2},
pages = {53--70},
year = {2009},
doi = {10.1080/08990220701318163}}

@INPROCEEDINGS{vardar219_fingertip,
  author={Vardar, Yasemin and Wallraven, Christian and Kuchenbecker, Katherine J.},
  booktitle={2019 IEEE World Haptics Conference (WHC)}, 
  title={Fingertip Interaction Metrics Correlate with Visual and Haptic Perception of Real Surfaces}, 
  year={2019},
  volume={},
  number={},
  pages={395-400},
  doi={10.1109/WHC.2019.8816095}}

@article{skedung2020_finishing,
author = {Skedung, L and Harris, K.L. and Collier, E. S. and Rutland, M.W.},
title = {The finishing touches: the role of friction and roughness in haptic perception of surface coatings},
journal = {Experimental Brain Research},
volume = {238},
pages = {1511--1524},
year = {2020},
doi = {10.1007/s00221-020-05831-w}}

@article{callier2015_kinematics,
author = {Callier, Thierri and Saal, Hannes P. and Davis-Berg, Elizabeth C. and Bensmaia, Sliman J.},
title = {Kinematics of unconstrained tactile texture exploration},
journal = {Journal of Neurophysiology},
volume = {113},
number = {7},
pages = {3013-3020},
year = {2015},
doi = {10.1152/jn.00703.2014},
    note ={PMID: 25744883},
URL = {https://doi.org/10.1152/jn.00703.2014},
eprint = {https://doi.org/10.1152/jn.00703.2014}}

@INPROCEEDINGS{priyadarshini2019_perceptnet,
  author={K, Priyadarshini and Chaudhuri, Siddhartha and Chaudhuri, Subhasis},
  booktitle={2019 IEEE World Haptics Conference (WHC)}, 
  title={PerceptNet: Learning Perceptual Similarity of Haptic Textures in Presence of Unorderable Triplets}, 
  year={2019},
  volume={},
  number={},
  pages={163-168},
  doi={10.1109/WHC.2019.8816141}}

@article{BERGMANNTIEST2006_analysis,
title = {Analysis of haptic perception of materials by multidimensional scaling and physical measurements of roughness and compressibility},
journal = {Acta Psychologica},
volume = {121},
number = {1},
pages = {1-20},
year = {2006},
issn = {0001-6918},
doi = {https://doi.org/10.1016/j.actpsy.2005.04.005},
url = {https://www.sciencedirect.com/science/article/pii/S0001691805000570},
author = {Wouter M. {Bergmann Tiest} and Astrid M.L. Kappers}
}

@ARTICLE{hollins1993_multidimensional,
  author={Hollins, M and Faldowski, R and Rao, S and Young, F},
  journal={Perception $\&$ Pscyhophysics}, 
  title={Perceptual dimensions of tactile surface texture: A multidimensional scaling analysis}, 
  year={1993},
  volume={54},
  pages={697-705},
  doi={10.3758/BF03211795}}

@ARTICLE{drewing2018_systematic,
  author={Drewing, Knut and Weyel, Claire and Celebi, Hevi and Kaya, Dilan},
  journal={IEEE Transactions on Haptics}, 
  title={Systematic Relations between Affective and Sensory Material Dimensions in Touch}, 
  year={2018},
  volume={11},
  number={4},
  pages={611-622},
  doi={10.1109/TOH.2018.2836427}}

@ARTICLE{fishel2012_bayesien,
    AUTHOR={Fishel, Jeremy A. and Loeb, Gerald E.},       
TITLE={Bayesian Exploration for Intelligent Identification of Textures},
JOURNAL={Frontiers in Neurorobotics},
VOLUME={Volume 6 - 2012},
YEAR={2012},
URL={https://www.frontiersin.org/journals/neurorobotics/articles/10.3389/fnbot.2012.00004},
DOI={10.3389/fnbot.2012.00004},
ISSN={1662-5218}}

@ARTICLE{lim2025_machine,
  author={Lim, Chungman and Kim, Gyeongdeok and Kang, Su-Yeon and Seifi, Hasti and Park, Gunhyuk},
  journal={IEEE Transactions on Haptics}, 
  title={Can a Machine Feel Vibrations?: Predicting Roughness and Emotional Responses to Vibration Tactons via a Neural Network}, 
  year={2025},
  volume={18},
  number={3},
  pages={542-555},
  doi={10.1109/TOH.2025.3568804}}

@article{lederman1987_hand,
title = {Hand movements: A window into haptic object recognition},
journal = {Cognitive Psychology},
volume = {19},
number = {3},
pages = {342-368},
year = {1987},
issn = {0010-0285},
doi = {https://doi.org/10.1016/0010-0285(87)90008-9},
url = {https://www.sciencedirect.com/science/article/pii/0010028587900089},
author = {Susan J Lederman and Roberta L Klatzky}
}

@article{klatzky2025_action,
   author = "Klatzky, Roberta L.",
   title = "Haptic Perception and Its Relation to Action", 
   journal= "Annual Review of Psychology",
   year = "2025",
   volume = "76",
   number = "Volume 76, 2025",
   pages = "227-250",
   doi = "https://doi.org/10.1146/annurev-psych-011624-101129",
   url = "https://www.annualreviews.org/content/journals/10.1146/annurev-psych-011624-101129",
   publisher = "Annual Reviews",
   issn = "1545-2085",
   type = "Journal Article",
  }

@misc{vardar2025_multimodal,
      title={Multimodal Haptic Device Enabling Naturalistic Surface Rendering During Active Touch}, 
      author={Yasemin Vardar and Haewon Jeong and Khoa Anh},
      year={2025},
      url={https://doi.org/10.21203/rs.3.rs-7428140/v1}, 
}

@article{lepora,
author = {Nathan F. Lepora},
title ={Tactile robotics: Past and future},

journal = {The International Journal of Robotics Research},
volume = {0},
number = {0},
pages = {02783649261421615},
year = {2026},
doi = {10.1177/02783649261421615},
URL = {https://journals.sagepub.com/doi/abs/10.1177/02783649261421615},
eprint = {https://journals.sagepub.com/doi/pdf/10.1177/02783649261421615}}

@article{biswas2019_emerging,
author = {Biswas, Shantonu and Visell, Yon},
title = {Emerging Material Technologies for Haptics},
journal = {Advanced Materials Technologies},
volume = {4},
number = {4},
pages = {1900042},
doi = {https://doi-org.tudelft.idm.oclc.org/10.1002/admt.201900042},
url = {https://advanced-onlinelibrary-wiley-com.tudelft.idm.oclc.org/doi/abs/10.1002/admt.201900042},
eprint = {https://advanced-onlinelibrary-wiley-com.tudelft.idm.oclc.org/doi/pdf/10.1002/admt.201900042},
year = {2019}
}

@article{bermanntiest2010_tactual,
title = {Tactual perception of material properties},
journal = {Vision Research},
volume = {50},
number = {24},
pages = {2775-2782},
year = {2010},
note = {Perception and Action: Part I},
issn = {0042-6989},
doi = {https://doi.org/10.1016/j.visres.2010.10.005},
url = {https://www.sciencedirect.com/science/article/pii/S0042698910004967},
author = {Wouter M. {Bergmann Tiest}}
}

@ARTICLE{nam2020_stickiness,
AUTHOR={Nam, Saekwang  and Vardar, Yasemin  and Gueorguiev, David  and Kuchenbecker, Katherine J. },
TITLE={Physical Variables Underlying Tactile Stickiness During Fingerpad Detachment},
JOURNAL={Frontiers in Neuroscience},
VOLUME={Volume 14 - 2020},
YEAR={2020},
URL={https://www.frontiersin.org/journals/neuroscience/articles/10.3389/fnins.2020.00235},
DOI={10.3389/fnins.2020.00235},
ISSN={1662-453X}}

@ARTICLE{richardson2020_learningtopredict,
AUTHOR={Richardson, Benjamin A.  and Kuchenbecker, Katherine J. },       
TITLE={Learning to Predict Perceptual Distributions of Haptic Adjectives},
JOURNAL={Frontiers in Neurorobotics},
VOLUME={Volume 13 - 2019},
YEAR={2020},
URL={https://www.frontiersin.org/journals/neurorobotics/articles/10.3389/fnbot.2019.00116},
DOI={10.3389/fnbot.2019.00116},
ISSN={1662-5218}}

@article{filingeri2014_wetness,
author = {Filingeri, Davide and Fournet, Damien and Hodder, Simon and Havenith, George},
title = {Why wet feels wet? A neurophysiological model of human cutaneous wetness sensitivity},
journal = {Journal of Neurophysiology},
volume = {112},
number = {6},
pages = {1457-1469},
year = {2014},
doi = {10.1152/jn.00120.2014},
URL = {https://doi.org/10.1152/jn.00120.2014}}

@article{ho2017_thermalrecognition,
author = {Ho, Hsin-Ni},
title = {Material recognition based on thermal cues: Mechanisms and applications},
journal = {Temperature},
volume = {5},
number = {1},
pages = {36--55},
year = {2017},
doi = {10.1080/23328940.2017.1372042}}

@ARTICLE{Kodak2023,
  author={Kodak, Bence L. and Vardar, Yasemin},
  journal={IEEE/ASME Transactions on Mechatronics}, 
  title={FeelPen: A Haptic Stylus Displaying Multimodal Texture Feels on Touchscreens}, 
  year={2023},
  volume={28},
  number={5},
  pages={2930-2940},
  doi={10.1109/TMECH.2023.3264787}}

@ARTICLE{okamoto2013_dimensions,
  author={Okamoto, Shogo and Nagano, Hikaru and Yamada, Yoji},
  journal={IEEE Transactions on Haptics}, 
  title={Psychophysical Dimensions of Tactile Perception of Textures}, 
  year={2013},
  volume={6},
  number={1},
  pages={81-93},
  doi={10.1109/TOH.2012.32}}

@article{richardson2022_learn2feel,
title = "Learning to Feel Textures: Predicting Perceptual Similarities From Unconstrained Finger-Surface Interactions",
author = "Richardson, {Benjamin A.} and Yasemin Vardar and Christian Wallraven and Kuchenbecker, {Katherine J.}",
year = "2022",
doi = "10.1109/TOH.2022.3212701",
language = "English",
volume = "15",
pages = "705--717",
journal = "IEEE Transactions on Haptics",
issn = "1939-1412",
publisher = "IEEE",
number = "4",
}

@article{saga2020_classification,
author = {Satoshi Saga and Shotaro Agatsuma and Simona Vasilache and Shin Takahashi},
title = {Machine Learning-based Classification and Generation of Vibrotactile Information},
journal = {International Journal On Advances in Networks and Services},
volume = {13},
issue = {3,4},
pages = {115-124},
year = {2020}}

@INPROCEEDINGS{Zhang2024_visuotactile,
  author={Zhang, Peng and Du, Dongfeng and Jiang, Haoliang},
  booktitle={2024 IEEE 6th Advanced Information Management, Communicates, Electronic and Automation Control Conference (IMCEC)}, 
  title={Classification of Visual-Tactile Fusion Object Attributes based on the GAMP Attention Mechanism}, 
  year={2024},
  volume={6},
  number={},
  pages={1279-1282}}

@article{Zheng2016_deeplearning,
author = {Zheng, Haitian and Fang, Lu and Ji, Mengqi and Strese, Matti and Ozer, Yigitcan and Steinbach, Eckehard},
title = {Deep Learning for Surface Material Classification Using Haptic and Visual Information},
year = {2016},
issue_date = {December 2016},
publisher = {IEEE Press},
volume = {18},
number = {12},
issn = {1520-9210},
journal = {Trans. Multi.},
pages = {2407–2416},
numpages = {10}
}

@ARTICLE{Strese2020_exploratory,
  author={Strese, Matti and Brudermueller, Lara and Kirsch, Jonas and Steinbach, Eckehard},
  journal={IEEE Transactions on Haptics}, 
  title={Haptic Material Analysis and Classification Inspired by Human Exploratory Procedures}, 
  year={2020},
  volume={13},
  number={2},
  pages={404-424},
  doi={10.1109/TOH.2019.2952118}}

@conference{Shuvo2025_classification,
title = "Material Classification using Visio-Tactile Sensor for Haptic Feedback Generation",
author = "Shuvo, {Md Golam Rabby} and Sonya Coleman and Dermot Kerr and JP Quinn",
year = "2025",
month = feb,
day = "25",
language = "English",
pages = "1--20",
url = "https://robovis.scitevents.org/?y=2025",
}

@ARTICLE{Strese2017_multimodal,
  author={Strese, Matti and Schuwerk, Clemens and Iepure, Albert and Steinbach, Eckehard},
  journal={IEEE Transactions on Haptics}, 
  title={Multimodal Feature-Based Surface Material Classification}, 
  year={2017},
  volume={10},
  number={2},
  pages={226-239},
  doi={10.1109/TOH.2016.2625787}}

@ARTICLE{Devillard2025_database,
  author={Alexis W. M. Devillard and Aruna Ramasamy and Xiaoxiao Cheng and Damien Faux and Etienne Burdet},
  journal={Scientific Data}, 
  title={Tactile, Audio, and Visual Dataset During Bare Finger Interaction with Textured Surfaces}, 
  year={2025},
  volume={12},
  number={1},
  pages={484},
  issn = {2052-4463}}

@misc{Zackory2017_semisupervised,
      title={Semi-Supervised Haptic Material Recognition for Robots using Generative Adversarial Networks}, 
      author={Zackory Erickson and Sonia Chernova and Charles C. Kemp},
      year={2017},
      eprint={1707.02796},
      archivePrefix={arXiv},
      primaryClass={cs.RO},
      url={https://arxiv.org/abs/1707.02796}, 
}

@ARTICLE{chen2026_review,
  author={Z. Chen and Y. Huang and B. Zhang and D. Sun and X. Yu},
  journal={Nature Reviews Materials}, 
  title={Deformable materials and structures in wearable haptic interfaces.}, 
  year={2026},
  doi = {10.1038/s41578-025-00877-0}}

@inproceedings{balasubramanian2025_sens3,
title = "SENS3: Multisensory Database of Finger-Surface Interactions and Corresponding Sensations",
author = "Balasubramanian, {Jagan K.} and Kodak, {Bence L.} and Yasemin Vardar",
year = "2025",
doi = "10.1007/978-3-031-70058-3_21",
language = "English",
isbn = "978-3-031-70057-6",
volume = "1",
series = "Lecture Notes in Computer Science (including subseries Lecture Notes in Artificial Intelligence and Lecture Notes in Bioinformatics)",
publisher = "Springer",
pages = "262--277",
editor = "Hiroyuki Kajimoto and Pedro Lopes and Claudio Pacchierotti and Cagatay Basdogan and Monica Gori and Betty Lemaire-Semail and Maud Marchal",
booktitle = "Haptics",
url = "https://eurohaptics.org/ehc2024/",}

@article{Baum2013_visual,
  title={Visual and haptic representations of material properties.},
  author={Elisabeth Baumgartner and Christiane B. Wiebel and Karl R. Gegenfurtner},
  journal={Multisensory research},
  year={2013},
  volume={26 5},
  pages={429-55},
  url={https://api.semanticscholar.org/CorpusID:18983780}}

@techreport{Fix+Hodges:1951,
  author = "E. Fix and J. L. Hodges",
  year = "1951",
  title = "Discriminatory analysis---{N}onparametric discrimination: {C}onsistency properties",
  institution = "USAF School of Aviation Medicine",
  address = "Randolph Field, Texas",
  number = "21-49-004",
  month = feb
}

@article{Cover1967,
  title        = {Nearest neighbor pattern classification},
  author       = {Cover, Thomas M. and Hart, Peter E.},
  year         = {1967},
  journal      = {IEEE Transactions on Information Theory},
  volume       = {13},
  number       = {1},
  pages        = {21--27},
  doi          = {10.1109/TIT.1967.1053964},
  url          = {https://ieeexplore.ieee.org/document/1053964},
  note         = {CiteSeerX: 10.1.1.68.2616},
  addendum     = {S2CID: 5246200}
}

@article{SVM1,
  title        = {Support-vector networks},
  author       = {Cortes, Corinna and Vapnik, Vladimir},
  year         = {1995},
  month        = sep,
  journal      = {Machine Learning},
  volume       = {20},
  number       = {3},
  pages        = {273--297},
  doi          = {10.1007/BF00994018},
  url          = {https://doi.org/10.1007/BF00994018},
  issn         = {1573-0565},
}

@article{PhysRevE.111.045306,
  title = {Persistent homology for structural characterization in disordered systems},
  author = {Wang, An and Zou, Li},
  journal = {Phys. Rev. E},
  volume = {111},
  issue = {4},
  pages = {045306},
  numpages = {22},
  year = {2025},
  month = {Apr},
  publisher = {American Physical Society},
  doi = {10.1103/PhysRevE.111.045306},
  url = {https://link.aps.org/doi/10.1103/PhysRevE.111.045306}
}

@inproceedings{RF,
  title        = {Random Decision Forests},
  author       = {Ho, Tin Kam},
  year         = {1995},
  booktitle    = {Proceedings of the 3rd International Conference on Document Analysis and Recognition},
  address      = {Montreal, QC},
  date         = {1995-08-14/1995-08-16},
  pages        = {278--282},
  url          = {https://web.archive.org/web/20160417030218/http://ect.bell-labs.com/who/tkh/papers/random-forests.pdf},
  note         = {Archived from the original on 17 April 2016. Retrieved 5 June 2016}
}

@article{Ho1998,
  title        = {The Random Subspace Method for Constructing Decision Forests},
  author       = {Ho, Tin Kam},
  year         = {1998},
  journal      = {IEEE Transactions on Pattern Analysis and Machine Intelligence},
  volume       = {20},
  number       = {8},
  pages        = {832--844},
  doi          = {10.1109/34.709601},
  url          = {https://ieeexplore.ieee.org/document/709601},
  note         = {Bibcode: 1998ITPAM..20..832T},
  addendum     = {S2CID: 206420153}
}

@inbook{Hastie2009,
  title        = {10. Boosting and Additive Trees},
  author       = {Hastie, Trevor and Tibshirani, Robert and Friedman, Jerome H.},
  year         = {2009},
  booktitle    = {The Elements of Statistical Learning},
  edition      = {2},
  pages        = {337--384},
  publisher    = {Springer},
  address      = {New York},
  isbn         = {978-0-387-84857-0},
  url          = {https://web.archive.org/web/20091110212529/http://www-stat.stanford.edu/~tibs/ElemStatLearn/},
  note         = {Archived from the original on 10 November 2009}
}

@article{GB,
title = {Approximating XGBoost with an interpretable decision tree},
journal = {Information Sciences},
volume = {572},
pages = {522-542},
year = {2021},
issn = {0020-0255},
doi = {https://doi.org/10.1016/j.ins.2021.05.055},
url = {https://www.sciencedirect.com/science/article/pii/S0020025521005272},
author = {Omer Sagi and Lior Rokach}}

@article{Cybenko1989,
  title        = {Approximation by superpositions of a sigmoidal function},
  author       = {Cybenko, George},
  year         = {1989},
  journal      = {Mathematics of Control, Signals, and Systems},
  volume       = {2},
  number       = {4},
  pages        = {303--314},
  doi          = {10.1007/BF02551274},
  url          = {https://doi.org/10.1007/BF02551274},
  issn         = {1435-568X}
}

@article{NN,
  title = {Graph-based descriptors for condensed matter},
  author = {Wang, An and Sosso, Gabriele C.},
  journal = {Phys. Rev. E},
  volume = {111},
  issue = {6},
  pages = {064302},
  numpages = {43},
  year = {2025},
  month = {Jun},
  publisher = {American Physical Society},
  doi = {10.1103/PhysRevE.111.064302},
  url = {https://link.aps.org/doi/10.1103/PhysRevE.111.064302}
}

\section*{Acknowledgements}
This study was partially funded by the Dutch Research Council (NWO) under the VENI scheme (ID: 19153) and the European Research Council (ERC) under the Starting Grant scheme (ID: 101220242). 

\section*{Author contributions statement}
LZ: Conceptualization, Methodology, Investigation, Software, Formal Analysis, Data Curation, Visualization, Writing - original draft. YV: Conceptualization, Methodology, Visualization, Writing - original draft, Writing - review \& editing, Supervision, Resources, Project Administration, Funding Acquisition.

\section*{Data Availability Statement}The data and code that support the findings of this study will be publicly online upon acceptance.

\section*{Additional information}
\noindent\textbf{Competing interests}: Authors report no competing interests.

\section*{Figure Legends}
\textbf{Figure 1:} Conceptual schematic of the model framework and data preparation. A. Human material recognition and modeling approach. Humans perceive multisensory tactile signals that convey surface attributes, which are integrated to recognize materials. Our framework approximates this process using three models: Model 1 maps extracted features from tactile signals to psychophysical ratings; Model 2 translates these perceptual attributes into material classifications; and Model 3 directly associates tactile features with material categories, serving as a baseline for artificial intelligence classification.
B. Data preparation. Tactile signals and psychophysical sensation ratings were collected from participants interacting with natural surfaces in the SENS3 database~\cite{balasubramanian2025_sens3} using sliding, static contact, and pressing exploratory procedures.\\

\noindent\textbf{Figure 2:} Results for interpreting the initial stage of human tactile processing, in which features are mapped to perceptual attributes (Model 1). (A) The performance of the regression models, evaluated using Mean Squared Error (MSE) and R-squared ($R^2$) scores. A lower MSE and a higher $R^2$ indicate better regression performance. All results were obtained using Random Forest regression, except for the null model, which serves as a baseline. (B) Spearman correlation between all pairs of sensation ratings. (C) Illustration of multidimensional scaling (MDS) for adjective pairs. In each subplot, each dot represents a participant, and the distance between any two dots reflects the difference in their ratings across the 50 textures.  \\

\noindent\textbf{Figure 3:} Results comparing models of higher-level human cognitive processing (Model 2) and direct AI-based classification (Model 3). 
(A) Classification accuracy for material identification using sensory attributes data, pressing data, static contact, sliding, and a combined dataset (pressing, static contact, and sliding). 
(B) Scatter plot of all materials in two-dimensional perceptual space derived from Principle Component Analysis (PCA), which each point representing a material positioned according to its first and second principle component scores. 
(C) Feature importance scores for all sensation ratings computed using a Random Forest model trained to classify material classes from sensory attributes. 
(D) Scatter plot of all materials in two-dimensional feature space obtained from PCA. 
(E) Classification accuracy as a function of the number of top-ranked features selected by Random Forest, using the combined dataset (pressing, static contact, and sliding) for material classification.

\section*{Supplementary Information}
Supplementary Figure S1\\
Supplementary Figure S2\\
Supplementary Figure S3\\
Supplementary Table S1\\
Supplementary Table S2

\end{document}